\documentclass[10pt,letterpaper]{article}

\usepackage{cite}
\usepackage{amsmath,amssymb,amsfonts}
\usepackage{algorithmic}
\usepackage{graphicx}
\usepackage{textcomp}
\usepackage{xcolor}
\usepackage{url}
\usepackage{balance} 
\usepackage{hyperref}

\usepackage{longtable}

\def\BibTeX{{\rm B\kern-.05em{\sc i\kern-.025em b}\kern-.08em
    T\kern-.1667em\lower.7ex\hbox{E}\kern-.125emX}}
\begin{document}

\vspace*{0.15in}

\begin{flushleft}
{\Large Prompting Large Language Models for Supporting the Differential Diagnosis of Anemia
\textbf\newline{}
}
\smallskip

Elisa Castagnari\textsuperscript{1,2}, 
Lillian Muyama\textsuperscript{1,2}, 
Adrien Coulet\textsuperscript{1,2,$\star$}\\

\bigskip

\text{\textsuperscript{1}} Inria Paris, Paris, France
\\
\text{\textsuperscript{2}} Centre de Recherche des Cordeliers, Inserm, Université Paris Cité, Sorbonne Université, Paris, France
\\

\bigskip
\textsuperscript{$\star$} corresponding author: \texttt{adrien.coulet@inria.fr}

\end{flushleft}


\begin{abstract}
In practice, clinicians achieve a diagnosis by following a sequence of steps, such as laboratory exams, observations, or imaging. The pathways to reach diagnosis decisions are documented by guidelines authored by expert organizations, which guide clinicians to reach a correct diagnosis through these sequences of steps. While these guidelines are beneficial for following medical reasoning and consolidating medical knowledge, they have some drawbacks. They often fail to address patients with uncommon conditions due to their focus on the majority population, and are slow and costly to update, making them unsuitable for rapidly emerging diseases or new practices.
Inspired by clinical guidelines, our study aimed to develop pathways similar to those that can be obtained in clinical guidelines. 
We tested three Large Language Models (LLMs) — Generative Pre-trained Transformer 4 (GPT-4), Large Language Model Meta AI (LLaMA), and Mistral — on a synthetic yet realistic dataset to differentially diagnose anemia and its subtypes. By using advanced prompting techniques to enhance the decision-making process, we generated diagnostic pathways using these models. Experimental results indicate that LLMs hold huge potential in clinical pathway discovery from patient data, with GPT-4 exhibiting the best performance in all conducted experiments.
\end{abstract}


\section{Introduction}
For complex diagnostic decisions, clinicians typically follow diagnostic guidelines that outline the sequential steps necessary to reach a diagnosis. These guidelines, developed by panels of experts based on the best available evidence, aim to standardize and streamline clinical decisions through recommended procedures such as gathering information, making observations, and ordering laboratory tests. However, the development of guidelines is both costly and time-consuming, making it challenging to develop guidelines that comprehensively cover the entire spectrum of diseases. Therefore, there is a pressing need for more flexible and scalable methods to provide insights when clinical guidelines are incomplete or unavailable.

We believe that new techniques, trained on clinical data or embedding a large amount of domain knowledge, can complement traditional diagnostic guidelines. Our goal is to develop methods that assist the decision-making process in a step-by-step manner, as it has been importantly outlined in previous research \cite{b11}. Following such an approach has the potential to minimize unnecessary tests, optimize healthcare costs, and offer more personalized and accurate diagnoses, particularly for patients with uncommon conditions.

The extensive data available in Electronic Health Records (EHRs), as well as the emergence of large language models (LLMs) offers significant opportunities to enhance clinical practice. EHRs contain structured, semi-structured, and unstructured data about patients' health, including medications, laboratory test orders and results, diagnoses, and demographic information. Previous studies have leveraged EHRs to train machine learning (ML) methods to automatically suggest diagnoses for patients \cite{guo2024multi}. However, these studies typically use supervised ML methods to predict a single endpoint, represented as a class label.
For data-driven approaches to be truly adopted in clinical practice, diagnoses should not be limited to a single endpoint. Instead, they could be represented as a pathway that encompasses the steps of medical reasoning and decision-making. This work builds upon a previous study that employed Deep Reinforcement Learning (DRL) trained on EHRs to achieve this goal \cite{b7}. In this paper, we propose an approach that uses LLMs and explores different models, evaluate their performance on the basis of synthetic but realistic EHRs, and compare them with the DRL approach.

Anemia, defined as a lower-than-normal amount of healthy red blood cells, was chosen as the clinical condition for this study for three reasons: its diagnosis is primarily based on a series of laboratory tests available in most EHRs; it is a common diagnosis, implying that the associated amount of data is sufficient to train ML models; and the differential diagnosis of anemia is frequently complex, making its guidance particularly useful.

We chose to use LLMs because, like the DRL approach, they can construct models that operate sequentially, passing through various steps, or following a human-readable chain of thought. We propose adapting prompts for LLMs to progressively build individualized pathways of observations to make before suggesting a diagnostic decision. For instance, in the anemia use case, a pathway would involve a sequence of laboratory test requests, with their results guiding the decision to request additional tests or make a diagnosis. We believe that these constructed pathways can complement clinical guidelines to aid practitioners in decision-making during the diagnosis process.

Our main contributions are:
\begin{enumerate}
\item \textbf{Developing and Evaluating Prompts:} We created various prompts that communicate the diagnostic task to the LLM to generate diagnostic pathways, and compared their performance.
\item \textbf{Evaluating LLM Performance:} We compared the performance of different LLMs in generating diagnostic pathways and diagnosing patients.
\item \textbf{Comparative Analysis:} We compared the performance of LLM-generated pathways with those generated by a concurrent DRL-based approach, to assess differences and improvements.
\end{enumerate}

\section{Related Works}
The discovery of clinical pathways from patient data has been extensively studied, with unsupervised learning methods such as clustering \cite{1, 3, 5, 30}, topic modeling \cite{63, 99, 135}, and particularly process mining \cite{61, 65, 139} being the most prevalent. 
Other works used supervised learning methods, such as decision trees \cite{42, 36} and neural networks \cite{49, 54}, to learn patient pathways from EHRs. Meanwhile, reinforcement learning has been actively applied in recent years to learn dynamic treatment regimens for patients \cite{93, 117, 140}. 
The application of Deep Reinforcement Learning for diagnosis pathways using EHRs represents another innovative approach. Yu et al. \cite{126} used DRL methods for cost-effective clinical tasks, including the prediction of Acute Kidney Injury. In \cite{b7}, Muyama et al. 
used EHR data to train DRL models for clinical diagnosis pathway generation, formulating the diagnosis process as a sequential decision-making problem within a Markov Decision Process (MDP) framework. 

Moreover, the recent emergence of LLMs has seen their application in a wide range of tasks, including text generation, language translation, chatbot development, among others. Their potential in clinical reasoning has gained significant attention, particularly for enhancing diagnostic accuracy and interpretability. Various studies have explored different frameworks and methodologies to integrate LLMs into clinical decision-making processes, aiming to address the unique challenges of the medical domain.
Traditionally, the assessment of LLMs has focused on multiple-choice questions. However, recent studies have shifted towards free-response clinical questions, showing the promise of newer LLMs like Generative Pre-trained Transformer 4 (GPT-4) in diagnosing complex clinical cases.
In \cite{b1}, Wu et al. developed a framework called In-Context Padding to guide LLMs' reasoning with medical knowledge seeds. This method involved extracting medical entities from clinical contexts, inferring relevant entities using a knowledge graph, and padding these knowledge seeds into prompts to guide the inference process of LLMs.
Similarly, Kwon et al. \cite{b5} used the Reasoning-Aware Diagnosis Framework, which leverages prompt-based learning to generate diagnostic rationales. They formulated clinical reasoning as a Clinical Chain-of-Thought (CoT), allowing LLMs to provide insights into patient data and the reasoning paths toward diagnoses. 
Moreover, Savage et al. \cite{b4} developed diagnostic reasoning prompts to evaluate the interpretability of LLMs, such as GPT-4, in medicine. 

In addition to these specific frameworks, LLMs like the Pathways Language Model (PaLM) and its instruction-tuned variant, Flan-PaLM, have shown state-of-the-art performance on various medical question datasets \cite{b3}. These models leverage a combination of prompting strategies and instruction prompt tuning to enhance comprehension, knowledge recall, and reasoning.

If these related works highlight the potential of LLMs in clinical reasoning and decision-making, the study presented in the paper aims to evaluate the effectiveness of various prompting techniques and incorporate domain knowledge to improve the performance and explainability of LLMs for diagnosing anemia, specifically using Electronic Health Records. It explores how different prompting strategies, including examples, domain knowledge rules, and Chain-of-Thought reasoning, impact the performance of different LLMs for this task. Also, it provides a comparative error analysis between LLMS and a DRL approach used in a previous study.

\section{Methods}

\subsection{The Models}
We explored the programmatic use of three different LLMs, based on the Transformer architecture described in \cite{vaswani2017attention} to generate anemia diagnosis pathways. All three LLMs have been pre-trained on a large corpus of texts. The LLMs we used are:

\subsubsection{Generative Pre-trained Transformer}
Developed by OpenAI \cite{openai}, GPT \cite{brown2020language} is an LLM that can be applied to various generative tasks. In this work, we used the fourth iteration in the GPT series, named GPT-4. A previous iteration (GPT-3.5) was initially considered but ultimately discarded because it consistently had a worse performance than GPT-4 on our preliminary tasks. This study uses GPT-4 Turbo.

\subsubsection{Large Language Model Meta AI (LLaMA)} 
LLaMA \cite{touvron2023LLaMA} is a family of LLMs, developed by Meta AI\cite{metaai}. This study uses LLaMA-3.

\subsubsection{Mistral}
Mistral is an LLM created by Mistral AI \cite{mistralai}. This study uses Mistral7B v0.3.

These models were applied to clinical reasoning tasks and adapted to the specific case of finding the optimal sequence of actions needed to achieve an accurate anemia diagnosis. 
We note that while Mistral and LLaMA are open-source, GPT is not.

\subsection{Prompting}
To achieve our objective, we tested the LLMs with various prompt engineering approaches to enhance the decision-making process and evaluate it over a synthetic, but realistic dataset.
The considered prompting approaches were:
\begin{itemize}
     \item \textbf{Providing a closed list of answers}: The model is prompted with the specification of the set of possible answers it could answer.
     \item \textbf{Setting a ``Persona'' \cite{b10}}: Setting a personality or character within the prompting, so that the LLM reasons and answers as if it was that personality. 
     \item \textbf{Providing examples: } Providing the LLM with one to $n$ problem-solution or input-output examples so it has a better understanding of the desired output.
    \item \textbf{Sequential prompting:} Instead of providing a single question and receiving a single answer, the model is prompted in a sequential manner, where at each step, it asks one question and gains new information about the patient, which in turn feeds the next step of the process.
    \item \textbf{Providing sets of rules:} Feeding the LLM with rules from existing clinical diagnosis guidelines, to guide its responses.
    \item \textbf{Chain-of-thought \cite{wei2022chain}}: The model is prompted to generate a step-by-step (chain) reasoning to explain its response, which breaks down the task into smaller reasoning steps.
\end{itemize}




Our various prompting experiments fall in two major categories: the \emph{Plain prompt}, where we provide the LLM with all the patient's features (\textit{e.g.} lab test results) at once and the LLM responds once with a diagnosis; and the \emph{Sequential prompt}, where the LLM requests for patient features one-by-one, and values are provided by our program to the LLM until a diagnosis is reached. 


\subsubsection{Plain Prompt} \hfill \break
Following prompting good practices and preliminary testing, the very first prompt we tested included the set of possible diagnoses, \textit{i.e.}, the set of possible answers.
To improve our results, we experimented with changing the persona of the models. We tested three persona settings: no persona, an AI Assistant specialized in anemia, and a clinician, expert in anemia. Based on preliminary and unreported empirical results, we employed the ``clinician'' persona in all our reported experiments.
Similarly, we explored in preliminary phases the impact of providing the LLM with examples, referred to as ``shots'', with three configurations: 0-shot, 1-shot, and few-shot (i.e., 3 examples). Based on the unreported empirical findings, we chose the 1-shot approach for the rest of our experiments. 

\begin{figure}
    \centering
    \includegraphics[width=1.0\linewidth]{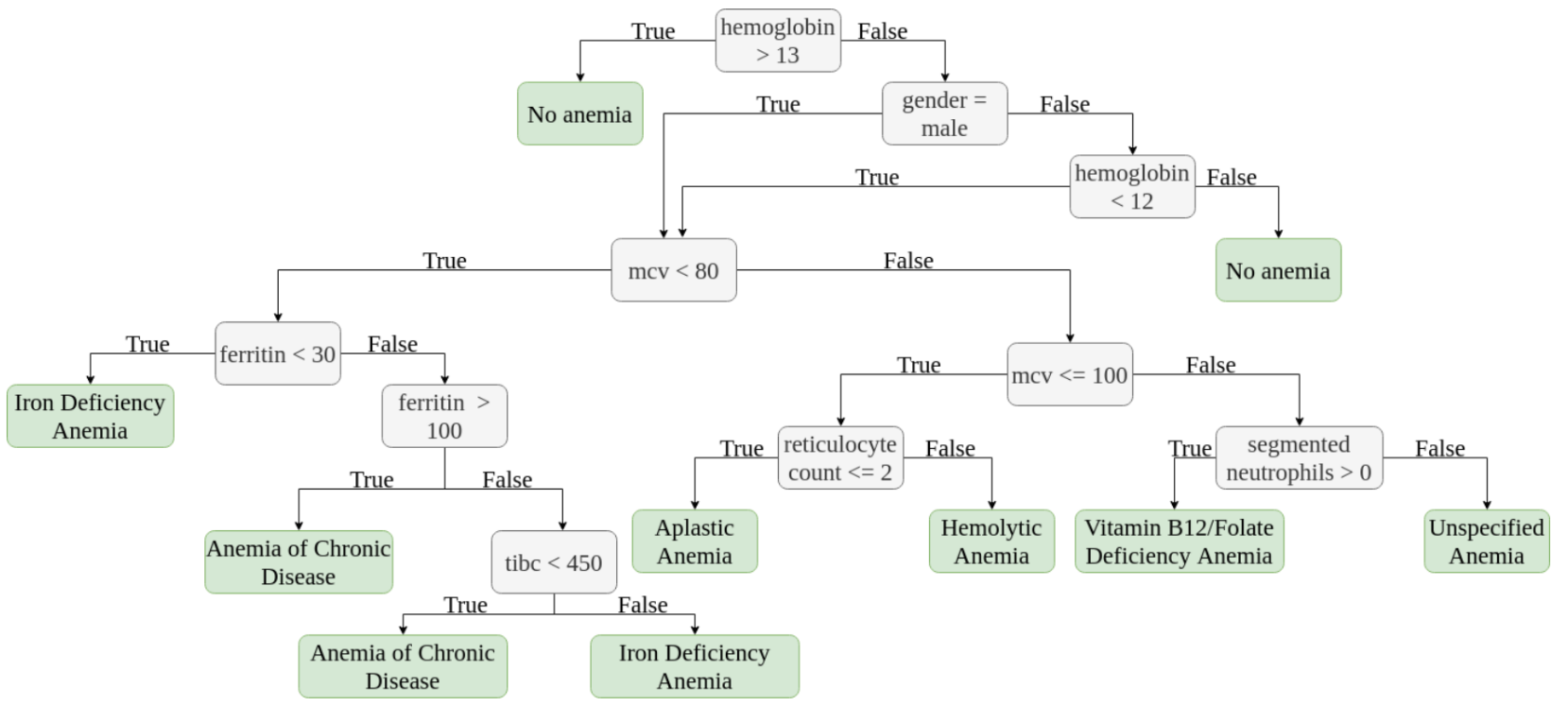}
    \caption{Anemia Decision Tree}
    \label{fig:anemia_dt}
\end{figure}

To assess the impact of incorporating additional domain knowledge into the prompt, we used a decision tree made from clinical diagnosis guidelines \cite{b7} and shown in Figure \ref{fig:anemia_dt}. We used it to provide LLMs with rules usually used for the differential diagnosis of anemia. 
We propose a pattern structure for translating pieces of the decision tree in natural language, as illustrated by the following prompt. \hfill 
\break
    
\noindent \textit{You are a clinician who is skilled in assessing whether a patient has anemia or not, and what type of anemia they have. You make the diagnosis based on their gender and laboratory test results. In your clinician role, you will give me the name of every lab test you took into consideration to determine the final diagnosis and the final diagnosis at the end.\\
Usually, you make a diagnosis based on the following rules:
\begin{itemize}
    \item [1)] Look for the \#\# value first.
    \item [2)] If the \#\# value is ** than ++, the diagnosis is @. If the \#\# value is ** than ++ but ** than ++, look for the \#\# of the patient.
    \item [] {\normalfont [If there are more than 2 cases:]}
    \item [3)] Otherwise, look for the \#\#. Here you can distinguish the following cases (named a, b,...,n):
    \begin{itemize}
        \item [a)] If the \#\# results are unavailable, the diagnosis is **.
        \item [b)] If the \#\# is ** than ++, look for the \#\# value. If the \#\# results are unavailable, the diagnosis is **. If the \#\# value is ** than ++, the diagnosis is @.
        \item [c)] ...
        \item [d)] {\normalfont [do step b]}
    \end{itemize}
\end{itemize}
For example, if you have that \#\#: **, \#\#: **, \#\#: **, the diagnosis will be @.} \\

\noindent where: 
\begin{itemize}
    \item \#\# is a laboratory test or the gender;
    \item ** is an operator (e.g., $<$, \(\leq\), $>$, \(\geq\), =);
    \item ++ is a numerical value;
    \item @ is a diagnosis type.
\end{itemize}
Please note that \#\#, **, ++ and @ are local variables and for this reason they can take various values within a single prompt.

Lastly, we evaluate the effectiveness of incorporating the chain-of-thought prompting strategy by asking the LLM to explain the steps leading to its response.

\subsubsection{Sequential Prompt}  \hfill \break
The plain prompt provides all the patient's information at once, effectively treating the diagnosis task as the prediction of a single endpoint without generating a pathway to reach that diagnosis.
Conversely, we consider a ``sequential'' prompt, where the LLM is instructed to request one patient feature at a time in each turn of the dialogue. The LLM receives results for each requested feature until it reaches a diagnosis. In this mode, at any given time, the LLM only has access to the information it has already specifically requested.

We restricted the models to inquire only about the features present in the decision tree and also specified the set of possible anemia diagnoses from which the model would select the final diagnosis.
Similar to the experiments with the plain prompt, we propose to evaluate the impact of providing of examples, incorporating rules from the decision tree, and the use of the chain-of-thought strategy.

\subsection{Dataset}
The experiments were conducted using the synthetic anemia dataset described in \cite{b7}, which was constructed based on the decision tree shown in Figure \ref{fig:anemia_dt}.
This dataset includes 17 features—\textit{hemoglobin, gender, mean corpuscular volume (MCV), ferritin, reticulocyte count,
segmented neutrophils, Total Iron Binding Capacity (TIBC), hematocrit, transferrin saturation (TSAT), red blood cells (RBC), serum iron, folate, creatinine, cholesterol, copper, ethanol,} and \textit{glucose}—and 8 classes: \textit{No anemia, Vitamin B12/Folate deficiency anemia, Unspecified anemia, Anemia of chronic disease (ACD), Iron deficiency anemia (IDA), Hemolytic anemia, Aplastic anemia,} and \textit{Inconclusive diagnosis}.

\begin{figure}[htbp]
    \centering
    \includegraphics[width=\linewidth]{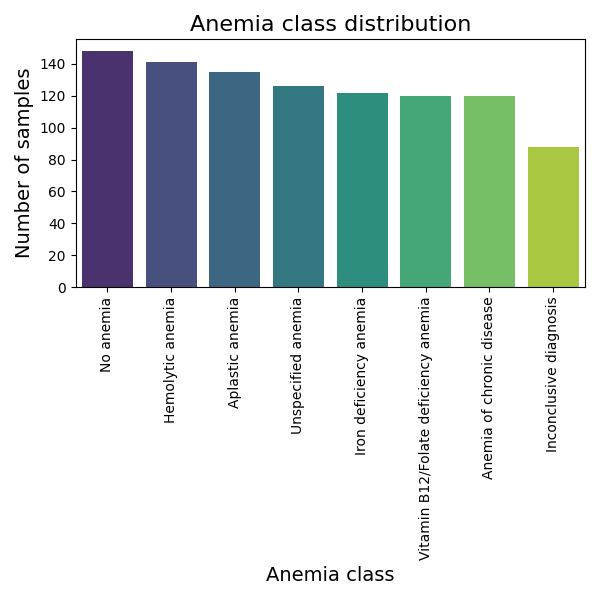}
    \caption{Distribution of patients across anemia classes.}
    \label{fig:class_distribution}
\end{figure}

For each diagnostic class, feature values were generated using a uniform probability distribution, with minimum and maximum values determined through a manual review of medical literature and thresholds from the decision tree. The dataset encompasses 70,000 patients.
A more detailed description of the dataset synthesis can be found in \cite{b7}.
For our experiments, we used 1,000 patients from it for both LLaMA and Mistral, whose class distribution is shown in Fig. \ref{fig:class_distribution}. We only used 250 patients for GPT-4 due to resource constraints.
We used the first 1,000 and 250 patients from the dataset for LLaMA/Mistral and GPT-4, respectively.

\subsection{Evaluation Approach and Implementation}
The performance was evaluated using the following metrics:
\begin{enumerate}
\item \textbf{Accuracy:} The proportion of patients that were correctly diagnosed.
\item \textbf{Mean Pathway Length:} The average number of actions in the diagnostic pathways generated by the model. 
\item \textbf{F1 Score:} The harmonic mean of the precision and recall scores. We report one-vs-rest and macro-averaging F1 scores.
\item \textbf{ROC-AUC (Receiver Operating Characteristic - Area Under Curve) Score:} This measures the model's ability to distinguish between different classes. We report one-vs-rest and macro-averaging ROC-AUC scores.
\item \textbf{Time:} The time for the LLMs to infer the results. 
\end{enumerate}

As mentioned, we conducted experiments on 1,000 patients, except for GPT-4, where we used 250 patients due to resource constraints. Due to inconclusive results with the seed parameter, our reported results are based on a single experiment run corresponding to the state of our machine at the time of experimentation.

The LLaMA and Mistral experiments were implemented using the Langchain Python library, and the GPT-4 experiments were conducted via the OpenAI API. 
The source code of these experiments is available at: \texttt{\small\url{https://anonymous.4open.science/r/anemia_diag_with_llm-1C1A/}}.
The reference for all the prompts used is in: \url{https://anonymous.4open.science/r/anemia_diag_with_llm-1C1A/prompts.txt}

\section{Results}

\subsection{Plain Prompts}
The baseline experiment involved a plain prompt specifying the set of anemia classes. The results of this experiment are shown in Table \ref{tab:baseline}.
All the models exhibited poor performance, with Mistral performing the worst and GPT-4 the best.

\begin{table}[h]
\caption{Plain prompt with specified anemia classes.}
\centering
\begin{tabular}{|c|c|c|c|c|}
  \hline
  \textbf{LLM }& \textbf{Accuracy }& \textbf{F1-Score} & \textbf{ROC-AUC }& \textbf{Time} \\
  \hline
  LLaMA & 29.59 & 15.98 & 59.07 & 10m 2.4s \\
  \hline
  Mistral & 11.3 & 6.51 & 49.97 & 8m 34.1s \\
  \hline
  \textbf{GPT-4} & \textbf{40.0} & \textbf{29.40} & \textbf{67.74} & 4m 11.8s \\
  \hline
\end{tabular}
\label{tab:baseline}
\end{table}

Next, we enhanced the baseline prompt with a single example of a diagnosis, such as: \textit{For example, you have that hemoglobin: 10g/dL, mean corpuscular volume: 83fL, reticulocyte count: 1.6\%. The diagnosis will be Aplastic anemia.} Results in Table \ref{tab:one_shot} show that incorporating an example slightly improved the performance of all the models.

\begin{table}[h]
\caption{Plain prompt with specified anemia classes and 1-shot.}
\centering
\begin{tabular}{|c|c|c|c|c|}
  \hline
  \textbf{LLM }& \textbf{Accuracy }& \textbf{F1-Score} & \textbf{ROC-AUC }& \textbf{Time} \\
  \hline
  LLaMA & 30.03 & 16.35 & 59.33 & 10m 9.0s \\
  \hline
  Mistral & 15.7 & 12.93 & 51.96 & 8m 59.25s \\
  \hline
  \textbf{GPT-4} & \textbf{44.0} & \textbf{32.65} & \textbf{66.35} & 4m 16.6s \\
  \hline
\end{tabular}
\label{tab:one_shot}
\end{table}

To experiment with the addition of domain knowledge to the LLMs' decision-making process, we manually converted the decision tree in Figure \ref{fig:anemia_dt} into natural language rules and added these rules to the prompt. Various templates for these rules were tested and the most suitable one was retained. The results in Table \ref{tab:decision_tree} show that
there was a significant improvement for GPT-4, with both GPT-4 and Mistral doubling their scores. LLaMA also improved, though to a lesser extent.
Despite the use of rules, there were still numerous misdiagnoses, particularly with LLaMA and Mistral, whereas the decision tree alone would perform a perfect classification.

\begin{table}[h]
\caption{Plain prompt with specified anemia classes, 1-shot, and decision tree rules.}
\centering
\begin{tabular}{|c|c|c|c|c|}
  \hline
  \textbf{LLM }& \textbf{Accuracy }& \textbf{F1-Score} & \textbf{ROC-AUC }& \textbf{Time} \\
  \hline
  LLaMA & 36.1 & 28.07 & 62.89 & 10m 36.4s \\
  \hline
  Mistral & 30.7 & 26.32 & 59.44 & 18m 56.0s \\
  \hline
  \textbf{GPT-4} & \textbf{82.0} & \textbf{74.91} & \textbf{87.49} & 4m 14.6s \\
  \hline
\end{tabular}
\label{tab:decision_tree}
\end{table}

Finally, we added the Chain-of-Thought to the plain prompt. The results, displayed in Table \ref{tab:cot}, revealed a massive improvement for LLaMA and a reasonable improvement for GPT-4. Mistral's performance did not show any significant change. Also, this approach made the models' decisions more explainable, allowing us to identify the causes of  misdiagnoses and misunderstandings of the models. This enabled us to better understand the model's errors and areas of confusion.

\begin{table}[h]
\caption{Plain prompt with specified anemia classes, 1-shot, decision tree rules , and Chain-of-Thought}
\centering
\begin{tabular}{|c|c|c|c|c|}
  \hline
  \textbf{LLM} & \textbf{Accuracy} & \textbf{F1-Score} & \textbf{ROC-AUC} & \textbf{Time} \\
  \hline
  LLaMA & 72.5 & 72.56 & 84.29 & 65m 47.6s \\
  \hline
  Mistral & 29.29 & 26.27 & 59.96 & 406m 47.0s \\
  \hline
  \textbf{GPT-4} & \textbf{92.8} & \textbf{91.41} & \textbf{94.90} & 17m 14.8s \\
  \hline
\end{tabular}
\label{tab:cot}
\end{table}

\subsection{Sequential Prompts}

We reproduced the same evaluation schema for the sequential prompts, but for simplicity, we only report the results of the two best-performing variants of these prompts.
The first, whose results are shown in Table \ref{tab:sequential}, involved a sequential prompt with specified anemia classes, 1-shot, and rules.
\begin{table}[h]
\caption{Sequential prompt with specified anemia classes, 1-shot, and decision tree rules}
\centering
\resizebox{0.5\textwidth}{!}{
\begin{tabular}{|c|c|c|c|c|c|}
  \hline
  \textbf{LLM} & \textbf{Accuracy} & \textbf{F1-Score} & \textbf{ROC-AUC} & \textbf{Avg. Length} &\textbf{ Time} \\
  \hline
  LLaMA & 49.1 & 43.19 & 70.63 & 3.35 & 22m 1s \\
  \hline
  Mistral & 27.7 & 22.89 & 58.89 & 3.63 & 23m 3.1s \\
  \hline
  \textbf{GPT-4} & \textbf{74.0 }& \textbf{69.56} & \textbf{85.60} & \textbf{4.16} & 17m 3.9s \\
  \hline
\end{tabular}
}
\label{tab:sequential}
\end{table}

The second prompt is similar, but included the CoT. Its results, presented in Table \ref{tab:sequential_cot}, show that adding CoT improved performance for both GPT-4 and LLaMA, while Mistral did not show similar gains. GPT-4 in this scenario achieved the best results of all our experiments.
Here, for comparison, we also display results from the Deep Q-Network (DQN) approach  used in \cite{b7} to generate anemia diagnosis pathways. 

\begin{table}[h!]
\caption{Sequential prompt with specified anemia classes, 1-shot, decision tree rules, and Chain-of-Thought}
\centering
\resizebox{0.5\textwidth}{!}{
\begin{tabular}{|c|c|c|c|c|c|}
  \hline
  \textbf{LLM} & \textbf{Accuracy} & \textbf{F1-Score} & \textbf{ROC-AUC} & \textbf{Avg. Length} &\textbf{ Time} \\
  \hline
  LLaMA & 61.4 & 62.98 & 78.62 & 6.10 & 262m 1.0s \\
  \hline
  Mistral & 24.9 & 20.88 & 56.89 & 3.13 & 106m 9.7s \\
  \hline
  \textbf{GPT-4 }& \textbf{98.4} & \textbf{98.21} & \textbf{99.09} & \textbf{4.08} & \textbf{23m 49.0s} \\
  \hline
  DQN & 97.5 & 97.50 & 98.60 & 4.82 & 1.6s\\
  \hline
\end{tabular}
}

\label{tab:sequential_cot}
\end{table}

\section{Discussion}
\subsection{LLM Performance}
In all our experiments, GPT-4 consistently had the best performance, followed by LLaMA, with Mistral performing the worst in all scenarios.
As for the prompting, adding an example to the prompt improved results across all the models. However, the performance of all models was notably poor before the introduction of rules from diagnosis guidelines into the prompts. Therefore, while these models have been pre-trained on vast amounts of texts, including medical texts, for specific use cases, it may be beneficial to fine-tune the models using task-specific datasets if the goal is to learn only from new patient data without considering existing guidelines. Similarly, using Retrieval Augmented Generation (RAG) \cite{b9}, could enable the identification and consideration of narrative clinical guidelines to fine-tune the embeddings, potentially leading to improved results. 
Additionally, incorporating the CoT strategy further improved the results for GPT-4 and LLaMA. This is because breaking down the task into smaller reasoning steps has been shown to enhance the reasoning abilities of LLMs.
However, this was not the case with Mistral. Upon analyzing its responses, we found that Mistral did not consistently apply the CoT for some patients, even if prompted.
Overall, LLMs present a valuable opportunity. By including domain knowledge in the form of rules expressed in natural langage to their input, they can generate diagnostic decision pathways in a conversational context and assist in comparing existing guidelines to newly suggested clinical pathways. 

\subsection{Error Analysis}
We conducted a detailed error analysis for the results of both plain and sequential prompts with CoT.
For the plain prompt, the majority of the misdiagnoses across all models were due to ``comparison errors'', where the LLMs were incorrect in determining whether a value was greater than, less than, or equal to another value.
For GPT-4, the remaining misdiagnoses were only attributed to the model not following the provided decision tree rules. 
For Mistral, a significant portion of the misdiagnoses (23.40\%) may have been caused by the model not applying the CoT, even though it was explicitly mentioned in the prompt. This explains why Mistral was the only model with a worse performance when CoT was prompted. 
Other misdiagnoses were due to diverse factors, such as not adhering to the decision tree rules, the model continuing the chat after a diagnosis was found (despite the conversation being supposed to terminate), and not accepting the provided value for a feature, among other issues.

In the sequential prompt, again, the majority of misdiagnoses were due to comparison errors, particularly for GPT-4, where these accounted for 100\% of the misdiagnoses. Specifically, for LLaMA, several \emph{hemolytic anemia} patients were diagnosed with \emph{aplastic anemia}, and vice versa. This is not surprising as both anemia types are found on the same branch of the decision tree as shown in Figure \ref{fig:anemia_dt}, with the value of a single feature \emph{(reticulocyte count)} making the difference.
Other errors were due to not following the decision tree rules provided, with Mistral requesting values for features that were not included in the dataset, despite the fact that the list of features to be selected was explicitly stated in the prompt.

\subsection{Generated Pathways}
\begin{figure}[htbp]
    \centering
    \includegraphics[width=\linewidth]{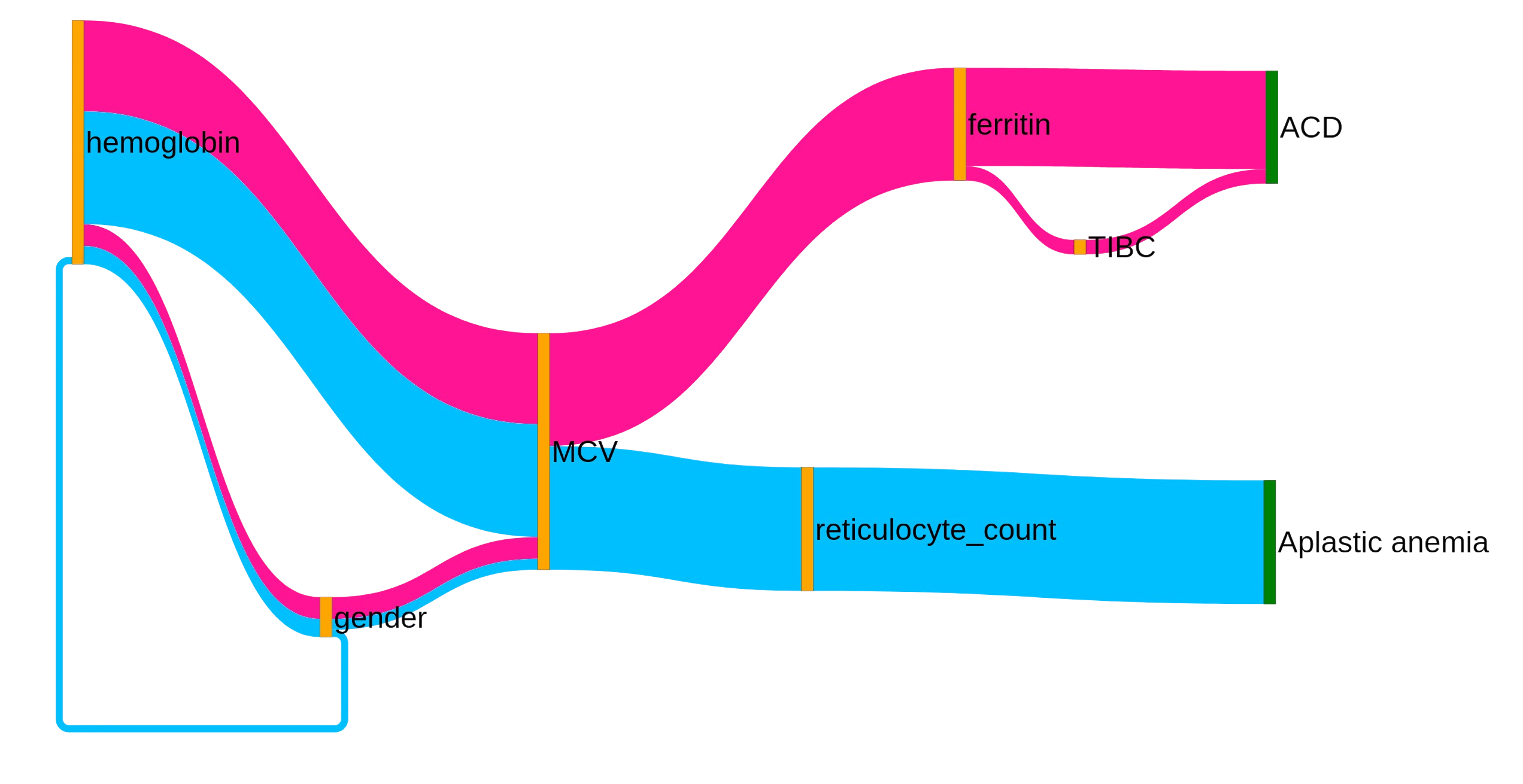}
    \caption{Pathways to Anemia of Chronic Disease (ACD) and Aplastic anemia, in pink and blue respectively, as generated by GPT-4.}
    \label{fig:sample_pathway}
\end{figure}

Fig. \ref{fig:sample_pathway} illustrates the pathways generated by GPT-4 using the sequential prompt and  CoT. In this Sankey diagram, the orange nodes represent the lab test results requested by the model in their order from left to right, while the dark green nodes represent the final diagnoses. The pink pathways correspond to patients diagnosed with ACD, and the blue pathways correspond to patients diagnosed with Aplastic anemia.

\begin{figure}[htbp]
    \centering
    \includegraphics[width=\linewidth]{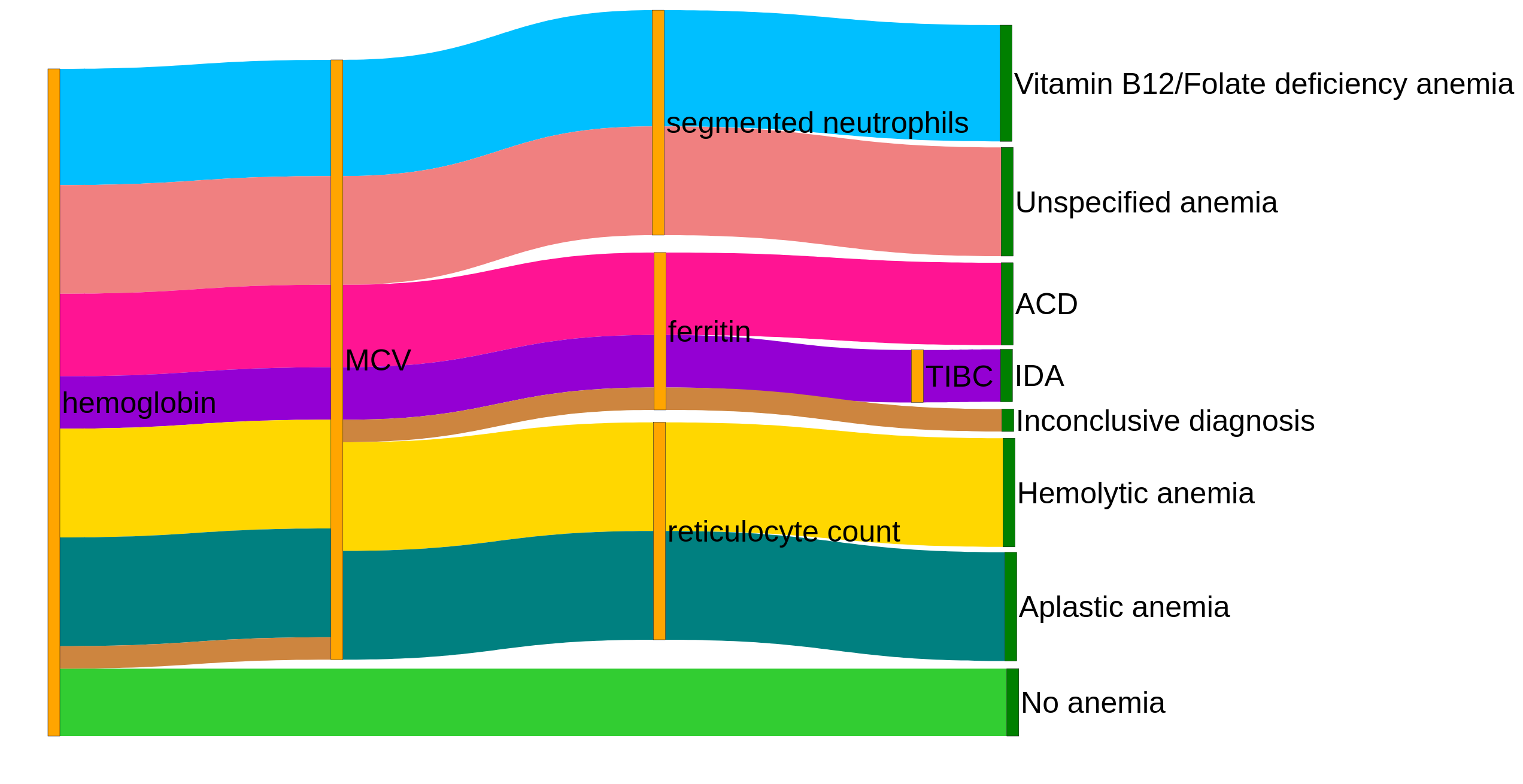}
    \caption{The commonest pathway to each anemia class for \emph{GPT-4}.}
    \label{fig:gpt_all}
\end{figure}

\begin{figure}[htbp]
    \centering
    \includegraphics[width=\linewidth]{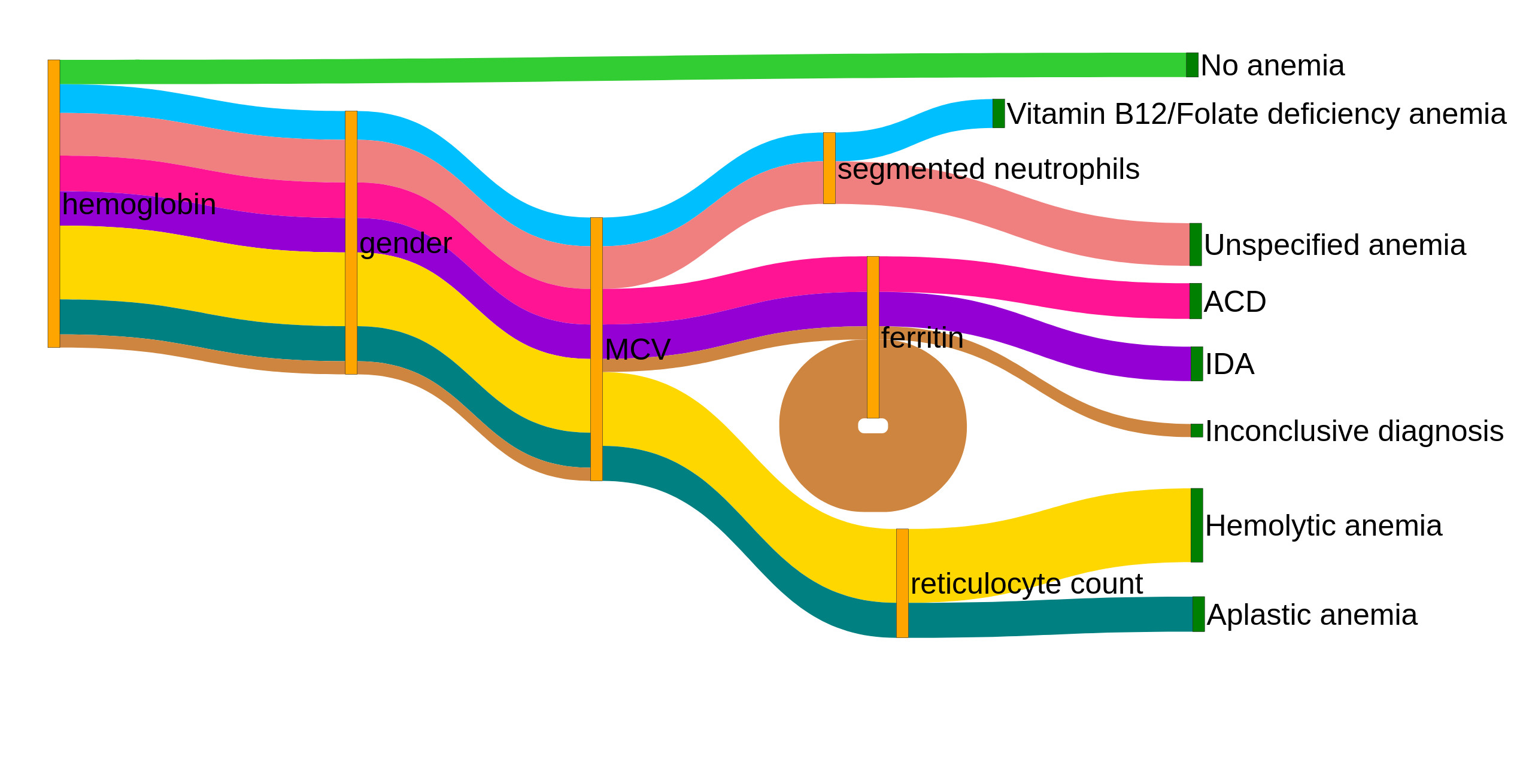}
    \caption{The commonest pathway to each anemia class for \emph{LLaMA}.}
    \label{fig:llama_all}
\end{figure}

\begin{figure}[!h]
    \centering
    \includegraphics[width=\linewidth]{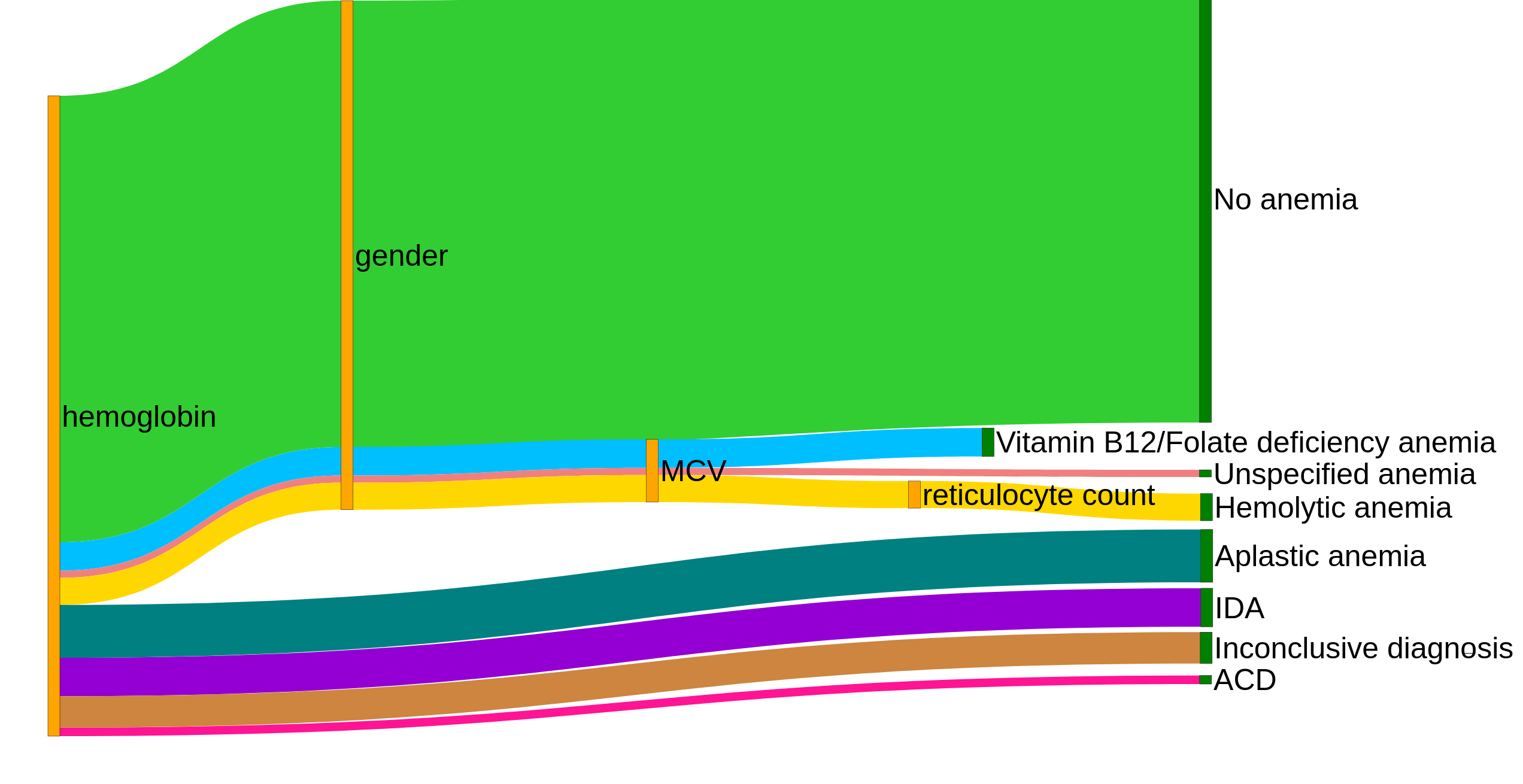}
    \caption{The commonest pathway to each anemia class for \emph{Mistral}.}
    \label{fig:mistral_all}
\end{figure}

Figs. \ref{fig:gpt_all}, \ref{fig:llama_all}, and \ref{fig:mistral_all} display the most common diagnostic pathways for each anemia class, as produced by GPT-4, LLaMA, and Mistral, respectively. Each class is represented by a different color, \textit{e.g.} pink for \emph{ACD}, green for \emph{No anemia}, and blue for \emph{Vitamin B12/Folate deficiency anemia}.
As shown in Fig. \ref{fig:sample_pathway} and \ref{fig:gpt_all}, the pathways generated by GPT-4 align closely with the decision tree used to label the dataset, whose rules were provided to the LLMs. 
For LLaMA, as illustrated in Fig. \ref{fig:llama_all}, although its most common pathways generally adhere to the decision tree rules, there were instances where it repeatedly requested the ferritin value before terminating with an inconclusive diagnosis. Additionally, unlike GPT-4, gender is a significant feature in almost all its most common pathways for each anemia class. 
In contrast, Fig. \ref{fig:mistral_all} shows that Mistral frequently made anemia diagnoses based solely on hemoglobin level. While hemoglobin is sufficient to confirm the presence of anemia, it is inadequate for identifying the specific type of anemia. This further explains Mistral's poor performance. 


To further investigate the similarities between the pathways generated by each model, we calculated the Levenshtein distance between the most common pathways for each anemia type produced by the LLMs. For comparison, we also generated pathways based on the decision tree depicted in Fig. \ref{fig:anemia_dt}. We encoded the pathways into strings, where each feature in the pathway was represented by a single character, and these characters were concatenated to form a string representing the entire pathway.  We compared the most common pathway for each anemia type across models and calculated the average Levenshtein distances. The results revealed that GPT-4's most common pathways perfectly matched those of the decision tree, suggesting that GPT-4 is the most effective model for rule-based diagnostic techniques. Among the LLMs, GPT-4 and LLaMA exhibited the greatest similarity, with an average Levenshtein distance of 1.75, while GPT-4 and Mistral followed with an average distance of 1.88. Conversely, the pathways generated by Mistral and LLaMA had the greatest divergence, with a mean Levenshtein distance of 2.63.

\subsection{Comparison with DRL approach}
In a previous study by Muyama et al. \cite{b7}, they used a DRL approach to learn clinical diagnostic pathways from EHR data. Specifically, they employed various extensions of a Deep Q-Network to achieve this goal. We decided to compare our sequential prompt-based results to theirs, as both approaches involve sequential decision-making and share the same objective.
It is important to note that the DQN models were trained and tested exclusively on EHR data, allowing them to learn diagnostic pathways solely from the data, without prior knowledge. 
In contrast, the LLMs used in our study, in addition to their pre-training on texts, were also provided with domain knowledge in the form of rules to follow in order to make a diagnosis. 

Moreover, the DQN-based study used a broader set of features, some of which were unrelated to anemia diagnosis. In our LLM-based approach, the features to be queried are explicitly mentioned in the prompt and incorporated into the decision tree rules. Despite these differences, among the three LLMs evaluated, only GPT-4 outperformed the DRL approach.
When comparing the diagnostic pathways generated by both methods, we found that the pathways from the DQN were, on average, longer than those generated by GPT-4. This is likely due to GPT-4 being constrained by the decision tree rules, which limit the number of features considered. 

Additionally, certain features, such as RBC, were prominent in the DQN-generated pathways, but absent in the LLM-generated pathways, as they are not included in the decision tree.
To further assess pathway similarity, we measured the Levenshtein distance between the most common pathways for each anemia class generated by each approach.  We found that the pathways generated by GPT-4 had the highest similarity to those produced by the DQN, with an average Levenshtein distance of 2. In comparison, the average Levenshtein distances between the DQN-generated pathways and those generated by LLaMA and Mistral were 2.25 and 2.36, respectively. This indicates that the LLM-generated pathways were more similar to each other than to those generated by the DQN, with the exception of those generated by Mistral and LLaMA, which exhibited the lowest similarity between them.

In terms of time (thus energy) efficiency, the trained DQN model quickly generates the diagnostic pathway by following its learned optimal policy, taking significantly less time than the LLMs, which must process the prompt's information at each step.

\subsection{Limitations}
While our study demonstrates the potential of LLMs for clinical decision-making, particularly in the diagnostic process, it has several limitations, primarily the use of synthetic data. Given these preliminary results, our current effort is to evaluate our methods on real-world data. 

Another limitation is the size of the dataset. Although we believe that the dataset used is a good indicator of the models' performance given that no additional model training is involved, we restricted the expense of funding resources by limiting the dataset size for GPT-4 to only a quarter of what was used for LLaMA and Mistral. While this provides a useful performance indication in terms of patient pathways and diagmoses, it should be considered a limitation and should be adjusted accordingly, especially for time comparisons.

A critical issue encountered is the variability in results due to the management of seeds for variable initialisation. Our findings from using LLaMA with the LangChain library indicate inconsistent seed behavior, particularly in sequential communication sessions with chains of messages. This inconsistency manifests as different results with the same seed and across different seeds. Conversely, in non-sequential scenarios, results were more consistent with the same seed and similar results with different seeds. Nevertheless, seed management within the system can lead to alterations in execution times and disparate outcomes. Consequently, the seed's unreliability drived us to exclude it from our considerations, emphasizing that results are ultimately contingent on system dynamics.

Another limitation observed is with the Sequential prompt and CoT, which is associated with a higher length of inference, associated with a high complexity and length of the prompt itself. Longer and more complex prompts hinder the LLM’s ability to precisely follow instructions, leading to inconsistent results. Each of our runs with the CoT introduced variations, as the model navigates through different scenarios and attempts to cover all possible cases. We observed that this complexity made it challenging for the model to consistently adhere to the decision tree rules.

Additionally, while some deviations from the decision tree’s prescribed rules were noted, these deviations may not necessarily indicate errors. They could potentially represent valid alternative diagnostic pathways. To validate these alternative solutions, a crucial next step is to involve a physician who can assess whether these deviations are indeed correct and whether the LLMs are capable of identifying more efficient or accurate diagnostic pathways.

Finally, in our comparison with the DRL approach, we acknowledge that multiple runs were conducted, resulting in various models, some of which demonstrated better performance than the one used in our study. This particular DQN model was selected because it had accuracy closest to the mean of the models evaluated in that study.

\section{Conclusions and Future Work}

In this study, we used three LLMs, \textit{i.e.} GPT-4, LLaMA, and Mistral to generate diagnostic pathways for anemia.
We employed various prompting techniques and assessed their impact on the models' performance.
Additionally, we compared these LLMs both to each other and to a DRL approach used in a previous study with a similar objective.
Our findings indicate that, in certain scenarios, in particular when domain knowledge is provided in the form of decision rules to the LLMs, they can enhance the decision-making process. 

For future work, we plan to apply our approach to real-world data and expand the decision tree to include a wider range of anemia types. This expansion will not only provide a more comprehensive evaluation of the LLMs' diagnostic capabilities but also address a broader spectrum of clinical scenarios. Additionally, we intend to explore the applicability of this approach to other clinical conditions beyond anemia. Furthermore, we aim to improve the LLMs' performance by exploring techniques such as LLM fine-tuning, prompt tuning, and RAG.
\\




\section{Acknowledgement}{This works is supported by the Agence Nationale de la Recherche under the France 2030 program, reference ANR-22-PESN-0007 ShareFAIR, and ANR-22-PESN-0008 NEUROVASC.}

\balance
\bibliographystyle{unsrt}
\bibliography{main}

\begin{thebibliography}{10}

\bibitem{b11}
Julia Adler-Milstein, Jonathan~H. Chen, and Gurpreet Dhaliwal.
\newblock {Next-Generation Artificial Intelligence for Diagnosis: From Predicting Diagnostic Labels to “Wayfinding”}.
\newblock {\em JAMA}, 326(24):2467--2468, 12 2021.

\bibitem{guo2024multi}
Lin~Lawrence Guo, Jason Fries, Ethan Steinberg, Scott~Lanyon Fleming, Keith Morse, Catherine Aftandilian, Jose Posada, Nigam Shah, and Lillian Sung.
\newblock A multi-center study on the adaptability of a shared foundation model for electronic health records.
\newblock {\em NPJ Digital Medicine}, 7(1):171, 2024.

\bibitem{b7}
Lillian Muyama, Antoine Neuraz, and Adrien Coulet.
\newblock Deep reinforcement learning for personalized diagnostic decision pathways using electronic health records: A comparative study on anemia and systemic lupus erythematosus.
\newblock {\em arXiv preprint arXiv:2404.05913}, 2024.

\bibitem{1}
Anastasia~A Funkner, Aleksey~N Yakovlev, and Sergey~V Kovalchuk.
\newblock Data-driven modeling of clinical pathways using electronic health records.
\newblock {\em Procedia Comput. Sci.}, 121:835--842, 2017.

\bibitem{3}
Verena~Hokino Yamaguti, Alberto Freitas, Anderson~Chidi Apunike, Rui Pedro Charters~Lopes Rijo, Domingos Alves, and Antonio~Ruffino Netto.
\newblock Clinical pathways and hierarchical clustering for tuberculosis treatment outcome prediction.
\newblock {\em Procedia Comput. Sci.}, 219:1373--1379, 2023.

\bibitem{5}
Shusaku Tsumoto, Tomohiro Kimura, and Shoji Hirano.
\newblock Automated dual clustering for clinical pathway mining.
\newblock In {\em 2020 IEEE International Conference on Big Data (Big Data)}, pages 5387--5396. IEEE, 2020.

\bibitem{30}
Rom{\'e}o Baulain, J{\'e}r{\'e}my Jov{\'e}, Dunia Sakr, Marine Gross-Goupil, Magali Rouyer, Marius Puel, Patrick Blin, C{\'e}cile Droz-Perroteau, R{\'e}gis Lassalle, and Nicolas~H Thurin.
\newblock Clustering of prostate cancer healthcare pathways in the french national healthcare database.
\newblock {\em Cancer Innov.}, 2(1):52--64, 2023.

\bibitem{63}
Xiao Xu, Tao Jin, Zhijie Wei, Cheng Lv, and Jianmin Wang.
\newblock {TCPM}: topic-based clinical pathway mining.
\newblock In {\em 2016 IEEE first international conference on connected health: Applications, systems and engineering technologies (CHASE)}, pages 292--301. IEEE, 2016.

\bibitem{99}
Xiao Xu, Tao Jin, Zhijie Wei, Jianmin Wang, et~al.
\newblock Incorporating topic assignment constraint and topic correlation limitation into clinical goal discovering for clinical pathway mining.
\newblock {\em J. Healthc. Eng.}, 2017, 2017.

\bibitem{135}
Zhengxing Huang, Zhenxiao Ge, Wei Dong, Kunlun He, and Huilong Duan.
\newblock Probabilistic modeling personalized treatment pathways using electronic health records.
\newblock {\em J. Biomed. Inform.}, 86:33--48, 2018.

\bibitem{61}
Jochen De~Weerdt, Filip Caron, Jan Vanthienen, and Bart Baesens.
\newblock Getting a grasp on clinical pathway data: an approach based on process mining.
\newblock In {\em Emerging Trends in Knowledge Discovery and Data Mining: PAKDD 2012 International Workshops: DMHM, GeoDoc, 3Clust, and DSDM, Kuala Lumpur, Malaysia, May 29--June 1, 2012, Revised Selected Papers 16}, pages 22--35. Springer, 2013.

\bibitem{65}
Xiaojin Zhang and Songlin Chen.
\newblock Pathway identification via process mining for patients with multiple conditions.
\newblock In {\em 2012 IEEE international conference on industrial engineering and engineering management}, pages 1754--1758. IEEE, 2012.

\bibitem{139}
Alireza Bakhshi, Erfan Hassannayebi, and Amir~Hossein Sadeghi.
\newblock Optimizing sepsis care through heuristics methods in process mining: A trajectory analysis.
\newblock {\em Healthc. Anal.}, 3:100187, 2023.

\bibitem{42}
Nicole~M Zimmerman, David Ray, Nicole Princic, Meghan Moynihan, Callisia Clarke, and Alexandria Phan.
\newblock Exploration of machine learning techniques to examine the journey to neuroendocrine tumor diagnosis with real-world data.
\newblock {\em Future Oncol.}, 17(24):3217--3230, 2021.

\bibitem{36}
Diti Machnes-Maayan, Soad~Haj Yahia, Shirly Frizinsky, Ramit Maoz-Segal, Irena Offengenden, Ron~S Kenett, Mona~I Kidon, and Nancy Agmon-Levin.
\newblock A clinical pathway for the diagnosis of sesame allergy in children.
\newblock {\em World Allergy Organ. J.}, 15(11):100713, 2022.

\bibitem{49}
Xiangyang Ye, Qing~T Zeng, Julio~C Facelli, Diana~I Brixner, Mike Conway, and Bruce~E Bray.
\newblock Predicting optimal hypertension treatment pathways using recurrent neural networks.
\newblock {\em Int. J. Med. Inform.}, 139:104122, 2020.

\bibitem{54}
Xijie Lin, Yuan Li, Yonghui Xu, Wei Guo, Wei He, Honglu Zhang, Lizhen Cui, and Chunyan Miao.
\newblock Personalized clinical pathway recommendation via attention based pre-training.
\newblock In {\em 2021 IEEE International Conference on Bioinformatics and Biomedicine (BIBM)}, pages 980--987. IEEE, 2021.

\bibitem{93}
Ying Liu, Brent Logan, Ning Liu, Zhiyuan Xu, Jian Tang, and Yangzhi Wang.
\newblock Deep reinforcement learning for dynamic treatment regimes on medical registry data.
\newblock In {\em 2017 IEEE international conference on healthcare informatics (ICHI)}, pages 380--385. IEEE, 2017.

\bibitem{117}
Lu~Wang, Wei Zhang, Xiaofeng He, and Hongyuan Zha.
\newblock Supervised reinforcement learning with recurrent neural network for dynamic treatment recommendation.
\newblock In {\em Proceedings of the 24th ACM SIGKDD international conference on knowledge discovery \& data mining}, pages 2447--2456, 2018.

\bibitem{140}
Haihong Guo, Jiao Li, Hongyan Liu, and Jun He.
\newblock Learning dynamic treatment strategies for coronary heart diseases by artificial intelligence: real-world data-driven study.
\newblock {\em BMC Med. Inform. Decis. Mak.}, 22(1):1--16, 2022.

\bibitem{126}
Zheng Yu, Yikuan Li, Joseph Kim, Kaixuan Huang, Yuan Luo, and Mengdi Wang.
\newblock Deep reinforcement learning for cost-effective medical diagnosis.
\newblock {\em arXiv preprint arXiv:2302.10261}, 2023.

\bibitem{b1}
Jiageng Wu, Xian Wu, and Jie Yang.
\newblock Guiding clinical reasoning with large language models via knowledge seeds.
\newblock {\em arXiv preprint arXiv:2403.06609}, 2024.

\bibitem{b5}
Taeyoon Kwon, Kai Tzu-iunn Ong, Dongjin Kang, Seungjun Moon, Jeong~Ryong Lee, Dosik Hwang, Beomseok Sohn, Yongsik Sim, Dongha Lee, and Jinyoung Yeo.
\newblock Large language models are clinical reasoners: Reasoning-aware diagnosis framework with prompt-generated rationales.
\newblock In {\em Proceedings of the AAAI Conference on Artificial Intelligence}, volume~38, pages 18417--18425, 2024.

\bibitem{b4}
Thomas Savage, Ashwin Nayak, Robert Gallo, Ekanath Rangan, and Jonathan~H Chen.
\newblock Diagnostic reasoning prompts reveal the potential for large language model interpretability in medicine.
\newblock {\em NPJ Digital Medicine}, 7(1):20, 2024.

\bibitem{b3}
Karan Singhal, Shekoofeh Azizi, Tao Tu, S~Sara Mahdavi, Jason Wei, Hyung~Won Chung, Nathan Scales, Ajay Tanwani, Heather Cole-Lewis, Stephen Pfohl, et~al.
\newblock Large language models encode clinical knowledge.
\newblock {\em Nature}, 620(7972):172--180, 2023.

\bibitem{vaswani2017attention}
Ashish Vaswani, Noam Shazeer, Niki Parmar, Jakob Uszkoreit, Llion Jones, Aidan~N Gomez, {\L}ukasz Kaiser, and Illia Polosukhin.
\newblock Attention is all you need.
\newblock {\em Advances in neural information processing systems}, 30, 2017.

\bibitem{openai}
Open{AI}.
\newblock \url{https://openai.com/}. Accessed: 2024-08-06.

\bibitem{brown2020language}
Tom Brown, Benjamin Mann, Nick Ryder, Melanie Subbiah, Jared~D Kaplan, Prafulla Dhariwal, Arvind Neelakantan, Pranav Shyam, Girish Sastry, Amanda Askell, et~al.
\newblock Language models are few-shot learners.
\newblock {\em Advances in neural information processing systems}, 33:1877--1901, 2020.

\bibitem{touvron2023LLaMA}
Hugo Touvron, Thibaut Lavril, Gautier Izacard, Xavier Martinet, Marie-Anne Lachaux, Timoth{\'e}e Lacroix, Baptiste Rozi{\`e}re, Naman Goyal, Eric Hambro, Faisal Azhar, et~al.
\newblock Llama: Open and efficient foundation language models.
\newblock {\em arXiv preprint arXiv:2302.13971}, 2023.

\bibitem{metaai}
Meta {AI}.
\newblock \url{https://www.meta.ai/}. Accessed: 2024-08-06.

\bibitem{mistralai}
Mistral {AI}.
\newblock \url{https://www.mistral.ai/}. Accessed: 2024-08-06.

\bibitem{b10}
Jamil Zaghir, Marco Naguib, Mina Bjelogrlic, Aur{\'e}lie N{\'e}v{\'e}ol, Xavier Tannier, and Christian Lovis.
\newblock Prompt engineering paradigms for medical applications: scoping review and recommendations for better practices.
\newblock {\em arXiv preprint arXiv:2405.01249}, 2024.

\bibitem{wei2022chain}
Jason Wei, Xuezhi Wang, Dale Schuurmans, Maarten Bosma, Fei Xia, Ed~Chi, Quoc~V Le, Denny Zhou, et~al.
\newblock Chain-of-thought prompting elicits reasoning in large language models.
\newblock {\em Advances in neural information processing systems}, 35:24824--24837, 2022.

\bibitem{b9}
Patrick Lewis, Ethan Perez, Aleksandra Piktus, Fabio Petroni, Vladimir Karpukhin, Naman Goyal, Heinrich K{\"u}ttler, Mike Lewis, Wen-tau Yih, Tim Rockt{\"a}schel, et~al.
\newblock Retrieval-augmented generation for knowledge-intensive nlp tasks.
\newblock {\em Advances in Neural Information Processing Systems}, 33:9459--9474, 2020.

\end{thebibliography}

\end{document}